# GUN: Gradual Upsampling Network for Single Image Super-Resolution

**Yang Zhao[1, 2], Guoqing Li[1], Wenjun Xie[1, 2], Wei Jia[1, 2], Hai Min[1], and Xiaoping Liu[1, 2]**

[1]The School of Computer and Information, Hefei University of Technology, 193 Tunxi Road, Hefei 230011, China
[2] Anhui Province Key Laboratory of Industry Safety and Emergency Technology, Hefei, 230009, China

Corresponding author: Wenjun Xie (e-mail: wjxie@ hfut.edu.cn).

This work was partly supported by the grant of National Science Foundation of China 61602146, 61673157, fundamental research funds for the Central Universities of China JZ2017HGTB0189, the grant of National Science Foundation of China 61672063, 61702154, and 61402018.

**ABSTRACT** In this paper, an efficient super-resolution (SR) method based on deep convolutional neural network (CNN) is proposed, namely Gradual Upsampling Network (GUN). Recent CNN based SR methods often preliminarily magnify the low resolution (LR) input to high resolution (HR) and then reconstruct the HR input, or directly reconstruct the LR input and then recover the HR result at the last layer. The proposed GUN utilizes a gradual process instead of these two commonly used frameworks. The GUN consists of an input layer, multiple upsampling and convolutional layers, and an output layer. By means of the gradual process, the proposed network can simplify the direct SR problem to multistep easier upsampling tasks with very small magnification factor in each step. Furthermore, a gradual training strategy is presented for the GUN. In the proposed training process, an initial network can be easily trained with edge-like samples, and then the weights are gradually tuned with more complex samples. The GUN can recover fine and vivid results, and is easy to be trained. The experimental results on several image sets demonstrate the effectiveness of the proposed network.

## I. INTRODUCTION

Single image super-resolution (SISR), which is also known as image upsampling, upscaling, or magnification, is a classical problem in computer vision and image processing. Generally, the aim of SISR is to reconstruct a high-quality (HQ) and high-resolution (HR) image from a single low-resolution (LR) input. It is a typical ill-posed problem since the detailed information of LR image is lost. Although many important progresses have been made in the past several decades, how to recover a HQ and HR image with low cost is still a fundamental and challenging task.

The basic SISR method is interpolation-based algorithm, such as nearest-neighbor, bilinear, bicubic [1], [2]. Unfortunately, interpolation often causes blurring, jaggy, and ringing effects. Hence, many methods have been proposed to suppress these unnatural artifacts by means of different strategies, such as introducing edge prior knowledge [3]-[5], altering interpolated grid [6]-[9], and sharpening the edges [10]-[12], *etc*. These improved methods refine the unnatural artifacts, but they still cannot recover extra details.

Reconstruction-based algorithm is another type of classical SISR method. This kind of method is based on a fundamental constraint that the reconstructed HR image should be consistent with the original LR input. In order to reproduce better results, many extra constraints or image models have been proposed over the years, *e.g.*, gradient-based constraints [13]-[18], local texture constraint [19], total variation regularizer [20], [21], deblurring-based models [22]-[24], *etc*. However, the performance of these reconstruction-based algorithms often degrades rapidly when the magnification factor increases, because the basic similarity constraint is defined on the LR space.

Example-based or learning-based method has received increasing attention in recent years. This kind of algorithm tries to reconstruct the missing details via lots of known LR/HR example-pairs. Learning-based method is first presented in [25] and further developed in [26]-[50]. Many typical and effective learning-based models have been proposed, such as neighbor embedding based algorithms [26]-[29], sparse representation based methods [30]-[37], and local self-exemplar models [38]-[41]. Although these methods can recover sharp edges with fine details, the computation cost of them is often quite high. The most time-consuming process of these methods is the patch-by-patch optimization of representation coefficients or weights. Recently, some fast and high-performance SISR models have been presented, *i.e.*, anchor neighborhood regression methods [42], [43], and SISR forests [59], [66]. These models can obtain obvious speedup by means of pre-computed projection matrix or efficient random forest algorithm.

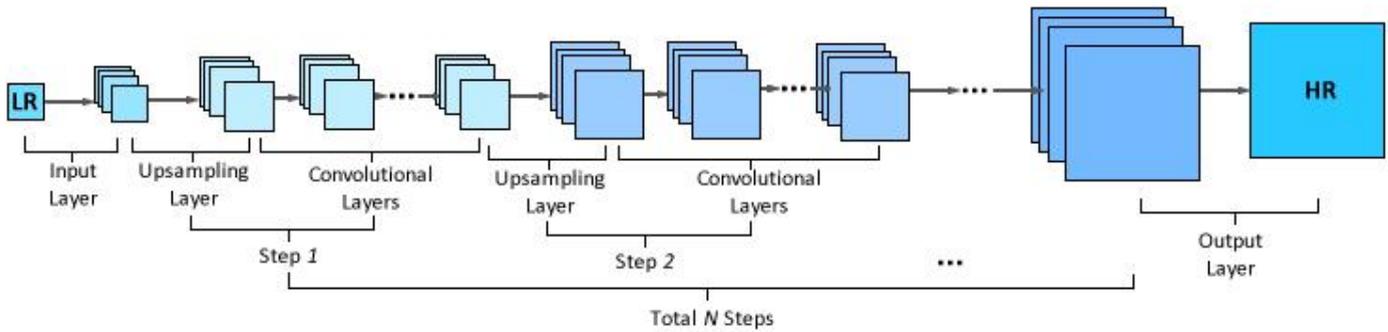

FIGURE 1. The framework of the proposed GUN

In recent several years, deep neural network (DNN) based methods have been widely applied in computer vision tasks and have achieved impressive results. Many DNN based SISR algorithms have also been proposed. Dong *et al.* [44] presented an effective SISR method by means of a shallow convolutional neural network (CNN). Kim *et al.* [45] further improve this method by successfully training a narrower and deeper CNN for SISR. Furthermore, many other DNN models have been applied in super-resolution (SR) scenario, such as deep residual network [55], sparse convolutional network [48], [60], recursive convolutional network [46], bidirectional recurrent convolutional network [54], collaborative local auto-encoder [53] and so on. In many CNN-based methods, the SISR is treated as an image reconstruction problem, and a general convolutional network without pooling and fully-connected layers is often used. The CNN has a strong ability to fit a highly nonlinear regression problem, and thus these CNN-based SR methods have achieved state-of-the-art results. The upsampling strategies in these CNN-based methods can be roughly divided into two categories: some methods preliminarily magnify the LR input to high resolution and then utilize the network to reconstruct the HR inputs [44]-[48]; some other methods directly reconstruct the LR input by means of convolutional networks and then reform the HR result in the last layer [49], [50]. Most recently, the generative adversarial network (GAN)-based SR methods [55], [56] reveal a possible way to recover fine texture details.

Most recently, many state-of-the-art deep residual network (ResNet)-based SR methods [51] have been proposed, and unceasingly refresh the PSNR records [52]. Owing to efficient residual connection and larger datasets, recent SR networks become increasingly deeper and better. However, deeper network also required more computational computation cost in many SR applications, such as FHD or UHD video reconstruction. This paper doesn't focus on training a very deep network to chasing the PSNR record, but tries to simplify traditional CNN-based SR model and make it easier to be trained. Concretely, a gradual upsampling network (GUN) is proposed, which gradually reconstructs the LR input to larger resolution by introducing several upsampling layers. Different to the methods [44] and [45], which directly reconstruct the interpolated HR input via deep network, the proposed GUN can be regarded as the concatenation of many sub-networks. In each sub-network, the target magnification factor is very small and thus the difficulty of learning in each step can be reduced. Gradual upsampling is a commonly used strategy in SR and other similar ill-posed problems. For example, direct magnifying with large factor is difficult in local self-exemplar methods, because merely finite self-examples can be used. Some methods [39] thus utilize the multistep magnification and adopt a very small factor (*e.g.*, $1.1 \times$, $1.2 \times$) in each step to reduce the difficulty of reconstruction. To take another example, it is also very hard to directly generate a HR image in generative model, the Laplacian pyramid GAN [57] and a CNN-based method [58] both enforce gradual upscaling process to yield a final full resolution output. Compared to the conventional CNN-based SR networks [44], [45], the proposed GUN has the following merits,

1) The gradual upsampling can relax the difficult direct magnification task to several easier upsampling problems with very small factor. The GUN is thus easier to be learned.

2) A gradual training method is presented for GUN. The gradual training process can rapidly train an initial network with finite simple samples. The GUN is then gradually optimized to reproduce HQ and HR results by continually adding more complicated samples to the training process.

3) Moreover, compared with some direct upsampling methods [44]-[48], the gradual upsampling can reduce the resolution of feature maps during the convolutional process, and therefore can decrease the computational cost.

The following paragraphs of this paper are organized as follows. Section II introduces the proposed GUN and gradual training process in details. Section III presents some implementation details. Experimental results are given in Section IV, and Section V concludes the paper.

## II. GRADUAL UPSAMPLING NETWORK

### A. THE PROPOSED NETWORK

As illustrated in Fig.1, the proposed GUN consists of several layers, *i.e.*, an input layer, multistep upsampling and convolutional layers, and a final output layer. In order to concisely illustrate the network, the activation and batch normalization (BN) layers are not shown in Fig.1. Each component of the GUN is described in the following.

#### 1) INPUT LAYER

Similar to the CNN-based SR methods [44], [45], the input layer of the GUN is also a typical convolutional (conv.) layer activated by the rectified linear units (ReLU). Hence, given a LR input $\boldsymbol{y}$ of size $(m_L, n_L)$, the output of the input layer is,

$$F_{in}(\boldsymbol{y}) = ReLU(\boldsymbol{\omega}_{in} * \boldsymbol{y} + \boldsymbol{b}_{in}). \quad (1)$$

In this paper, we use 64 filters with the size of $3 \times 3$ in the first layer, and thus the sizes of weights $\boldsymbol{\omega}_{in}$ and bias term $\boldsymbol{b}_{in}$ are $3 \times 3 \times 1 \times 64$ and $1 \times 64$, respectively. It should be noted that all the conv. processes in the GUN are with zero-padding, so that the resolution is invariant after the conv. process.

#### 2) UPSAMPLING LAYER

The upsampling layer can resize the input to a slightly larger resolution, which can be described as,

$$F_{up}^l(\boldsymbol{y}^l) = U \uparrow (\boldsymbol{y}^l), \quad (2)$$

where $\boldsymbol{y}^l$ is the input of the layer $l$, and $U \uparrow$ denotes the upsampling process. The magnification factors of $U \uparrow$ are mainly decided by the depth of the GUN, and are often much less than 2. In this paper, the traditional bicubic interpolation is adopted in the upsampling layers. It should be noticed that other interpolation methods or fast SR methods can also be applied in the upsampling layer. For a fair comparison with [44], [45], which adopt bicubic to magnify the original LR images, same bicubic interpolation is thus used in this paper. Correspondingly, the propagated error value passed by this layer need to be downsampled during the back propagation process:

$$\boldsymbol{\delta}^l = D \downarrow (\boldsymbol{\delta}^{l+1}), \quad (3)$$

where $\boldsymbol{\delta}^l$ denotes error value of the $l$-th layer, and $D \downarrow$ represents the downsampling process.

Suppose the size of final output HR image $\boldsymbol{x}$ is $(m_H, n_H)$, and the LR input $\boldsymbol{y}$ is gradually reconstructed with total $N$ steps, then the upsampled resolution of the $i$-th upsampling layer is computed as,

$$(m_L + i\Delta_m, n_L + i\Delta_n), \quad i = 1, 2, \cdots, N-1, \quad (4)$$

where,

$$\Delta_m = R\left(\frac{m_H - m_L}{N}\right),$$

$$\Delta_n = R\left(\frac{n_H - n_L}{N}\right),$$

$R(\cdot)$ denotes the round down function. Note that the upsampled resolution of the $N$-th step is fixed to $(m_H, n_H)$, so that the resolution of the final output is the same with the target.

Note that the de-convolutional (de-conv.) or un-pooling layer can also enlarge the resolution of the output. However, the un-pooling layer can merely upscale with a fixed integer factor, and the enlarged size of de-conv. layer is strictly determined by the size of filters or the stride of de-conv. process. Compared to un-pooling layer and de-conv. layer, the upsampling layer can magnify the input to a specified resolution more freely[1].

#### 3) CONVOLUTIONAL LAYERS

In the proposed GUN, the LR input is gradually upsampled within many steps. Each step contains an upsampling layer and several conv. layers. Motivated by the VGG-Net [61] and the VDSR [45], the stack of many $3 \times 3$ conv. layers can have an effective receptive field of larger sizes, *e.g.*, $5 \times 5$, $7 \times 7$, and so on. Therefore, we also set the size of filters to $3 \times 3$. However, the multiple $3 \times 3$ convolutions cannot represent $1 \times 1$ convolution. It has been proven that $1 \times 1$ convolution is valid in the SR problem [44]. We thus add a $1 \times 1$ conv. layer as the last layer in each step. Then the output of the conv. layer $l$ can be calculated as,

$$F^l(\boldsymbol{y}^l) = ReLU(\boldsymbol{\omega}^l * \boldsymbol{y}^l + \boldsymbol{b}^l). \quad (5)$$

We also utilize 64 feature maps in all the conv. layers. Hence, the size of $\boldsymbol{\omega}^l$ is $1 \times 1 \times 64 \times 64$ for the last conv. layer in each step, and the size of $\boldsymbol{\omega}^l$ for all the other conv. layers is $3 \times 3 \times 64 \times 64$. The size of $\boldsymbol{b}^l$ is therefore $1 \times 64$ for all the conv. layers. In addition, a BN layer is applied after each conv. layer in the GUN to enhance the capacity of the network.

#### 4) OUTPUT LAYER

The final output layer is computed as,

$$\boldsymbol{x} = \boldsymbol{\omega}_{out} * \boldsymbol{y}^l + \boldsymbol{b}_{out}, \quad (6)$$

---

[1] A gradually de-conv. network is also presented similar to the GUN. More details and related experimental results can be found in the supplementary materials: http://yzhaocv.weebly.com/projectpage/gun-supplementary

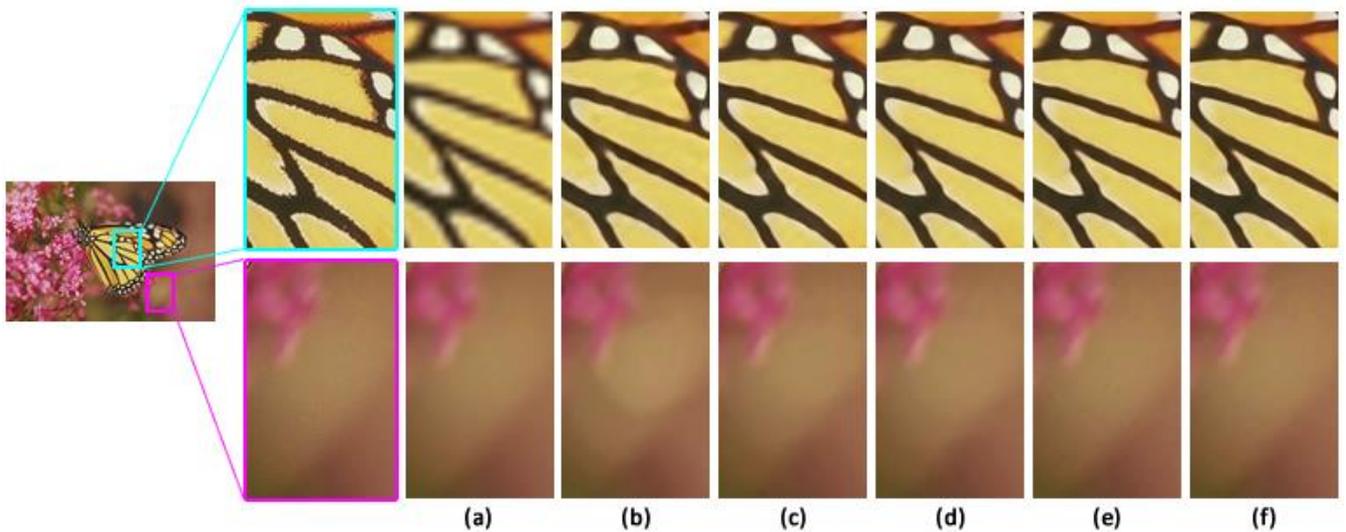

where the sizes of $\boldsymbol{\omega}_{out}$ and $\boldsymbol{b}_{out}$ are $3 \times 3 \times 64 \times 1$ and $1 \times 1$, respectively.

In the end of this sub-section, we further discuss the relationship between the GUN and some other gradual/cascaded SR methods. For instance, the original sparse coding based network (SCN) [67] is designed for fixed $2\times$ scale, an effective cascaded framework is thus introduced to achieve $4\times$ magnification by twice $2\times$ magnifications. This cascaded SR process is also proved to be better than direct $4\times$ SR. In [47] and [68], multiple $2\times$ upsampling processes are used in a cascade network to magnify image with large scales. These methods demonstrate the effectiveness of the cascaded framework, but the magnification factors of the proposed upsampling layers are much smaller than fixed $2\times$. These small factors in the GUN are similar to traditional local self-exemplar based methods [39] and cascaded collaborative local auto-encoder (CLA) method [53]. But different to [53], the proposed GUN fuses the gradual upsampling processes into an end-to-end network, while the CLA and self-exemplar blocks in [53] need to be optimized independently.

## B. TRAINING OF THE GUN

### 1) NORMAL TRAINING SETTINGS

As in other SR networks, the average mean squared error (MSE) is also used as the loss function for the GUN. The MSE loss can restrict the pixel-wise contents of the output are exactly consistent with that of the HR sample. Given a training image set $\{\boldsymbol{x}^n, \boldsymbol{y}^n\}_{n=1}^{N_s}$, the GUN can be trained by minimizing the following MSE loss function,

$$L(\theta) = \frac{1}{N_s} \sum_{n=1}^{N_s} \|f(\boldsymbol{y}^n; \theta) - \boldsymbol{x}^n\|^2, \tag{7}$$

where $f(\cdot)$ denotes the output of the network, and $\theta$ is the weight set of the GUN.

In this paper, we adopt the training image set proposed by Yang *et al.* [30], which is also used in many other learning-based methods [30]-[35], [42]-[45]. This training set contains 91 images downloaded from the internet. Many sample patches are randomly selected from each training image with overlapping, and these patches are further augmented by rotating with three orientations ($45°$, $90°$, and $180°$).

The proposed GUN is trained by utilizing mini-batch gradient descent based on backward propagation. Each mini-batch contains 64 image patches. The momentum parameter and weight decay are set as in [45]. The learning rate is initially set to $10^{-4}$ and then decreased by a factor of 10 after several epochs. We implement the GUN[2] by means of the MatConvNet[3] package [62].

### 2) GRADUAL TRAINING: FROM EASY TO DIFFICULT

Image set is very important to train an effective deep network. Related works often enlarge the quantity of training samples to refine the performance, *e.g.*, adding more images into the training set, extracting more patches from one image, and data augmentation (flipping, rotation, scaling, and so on). However, all these samples play the same role during the training process. In traditional learning-based SR methods, it can be found that the edge-like patterns with stable local structure are much easier to be learned in the dictionary than other kinds of patches. For example, most of the dictionary atoms learned via sparse representation or clustering are edge-like local patterns. Is it also easier to train the network by means of these edge-like patches? To answer this question, we adopt a gradual training process. In the gradual training, the patches which contain sharp edges are firstly chosen as the initial training set. Patches with flatter structure are then gradually added into the training set. By using the proposed training process, the GUN is firstly trained to magnify the sharp edges, and the details of sharp edge area are relatively easier to be learned. The network then learns to reconstruct more difficult situations by gradually fine-tuning the weights with more training samples.

---

[2] The demo codes of the GUN can be downloaded from the following website: http://yzhaocv.weebly.com/projectpage/gun

[3] http://www.vlfeat.org/matconvnet/

In this paper, the edge-like patches are selected by means of the average local gray value difference (ALGD), which can be computed as,

$$\mathrm{v}_{ALGD} = \frac{1}{N_p} \sum_{p=1}^{N_p} (g_p - \bar{g}), \qquad (8)$$

where $g_p$ $(p = 1,2,\cdots,N_p)$ denotes a pixel in an image patch, $N_p$ is the total number of pixels in that patch, and $\bar{g}$ denotes the average gray value of the patch. The edge-like patches can then be selected by comparing the ALGD value with the average ALGD value of the whole training set $(\overline{\mathrm{v}_{ALGD}})$ as follows,

$$\mathrm{v}_{ALGD} \geq \lambda \overline{\mathrm{v}_{ALGD}},$$

where $\lambda$ is an artificial parameters, and the $\lambda$ is orderly set as 1.2, 1, 0.8, 0.5, and 0 in our training process. At the first, the sharp-edge-patches with $(\lambda = 1.2)$ are utilized to form the initial training set. Take $4 \times$ magnification for example, merely about 29, 000 $12 \times 12$ patches are extracted. This training stage is convergence very fast within first 3-5 epochs, and it is much faster than the normal training process since the size of this initial training set is much smaller. The reconstructed results of different training stages are illustrated in Fig.2. From Fig.2(b) we can find that the GUN can learn to magnify sharp edges with faster training and less samples. But unfortunately, the flat areas are also over-sharpened. We then feed more patches with $(\lambda = 1)$ to the network, and the number of samples is increased to almost 82, 000 for $4 \times$ magnification. This training stage costs another 3 epochs. The upsampling results after this stage is shown in Fig.2(c). It can be found that the reconstructed edges become clearer, and the flat areas are much better than the former results. Similarly, the patches with $(\lambda = 0.8, 0.5, and\ 0)$ are added into the training set in turn, and finally over 300, 000 samples are utilized. By comparing the results in Fig.2, we can find that the gradual training process can reproduce both sharp edges and natural flat area by gradually tuning the weights. We implement each training stage about 3 epochs, and totally use 15-20 epochs to train

TABLE I. AVERAGE PSNR (dB) ON 'SET14' WITH DIFFERENT GRADUAL TRAINING STAGES

| $\lambda$ | $2 \times$ | $3 \times$ | $4 \times$ |
|---|---|---|---|
| Bicubic | 30.36 | 27.67 | 26.12 |
| 1.2 | 32.77 | 29.45 | 27.62 |
| 1 | 33.06 | 29.67 | 27.89 |
| 0.8 | 33.24 | 29.96 | 28.15 |
| 0.5 | 33.30 | 30.04 | 28.21 |
| 0 | 33.33 | 30.07 | 28.25 |

the GUN. This proposed training process roughly cost merely 2-3 hours on a PC using a Titan X GPU.

The average PSNR values of $2 \times$, $3 \times$, and $4 \times$ magnifications of different training stages are listed in the Table I. From which we can find that the initial training with small training set can already obtain fine PSNR results. These results are then slowly improved by tuning the network with more and complex training samples. Furthermore, the results of the gradual training also verify the prior knowledge about the training of SR network, i.e., the reconstruction of sharp edges with stable local structure are easier to be learned, and gradual training can make the network learn the SR task better, from easy to difficult.

Similar multistep training strategies have been successfully applied for many applications. For example, in a texture synthesis network [69], Li et al. observed the deep network has the following two properties, i.e., the network can learn one texture and then gradually learn other textures, and the network does not forget what is already learned. The proposed gradual training process for the GUN also partially verifies this opinion.

## C. DISCUSSION OF THE COMPUTATION COMPLEXITY

The computation complexity of the GUN can be computed as,

$$O\left\{(f_1^2 q_1)S_1 + \sum_{i=1}^{N} \left(\sum_{l=1}^{D} p_l f_l^2 q_l\right) S_i + (p_L f_L^2)S_L\right\} \qquad (9)$$

where $S_1$, $S_i$, and $S_L$ denote the size of the LR image, the size of the $i$-th step maps, and the size of the HR output, respectively. The $f_l$ is the filter size of the $l$-th layer, $p_l/q_l$ represents the number of input/output feature maps of the $l$-th layer, and $D$ denotes the depth of the sub-network in each step. It can be observed that the complexity is proportional to the size of image, the number of feature maps in each layer, filter size, and the depth of the network. As mentioned before, the $p_l/q_l$ is fixed as 64, and the filter size is set as small as 3 or 1 in the proposed network to avoid high computation complexity. Furthermore, the GUN has lower computation complexity than the direct upsampling network [44], [45], since the $S_i$ is smaller than the $S_L$.

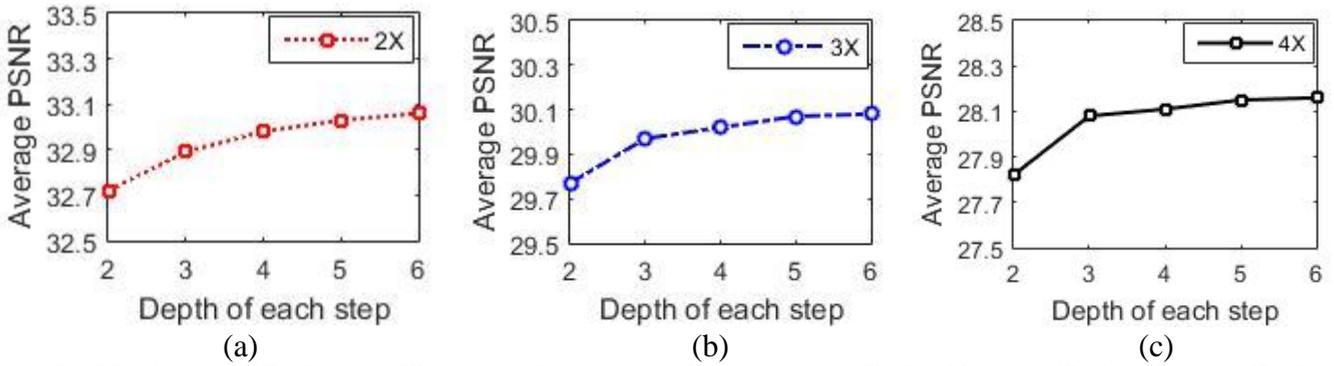

**FIGURE 3.** Average PSNR values with different depths in each step on 'Set14', (a) 2 × magnification, (b) 3 × magnification, (c) 4 × magnification.

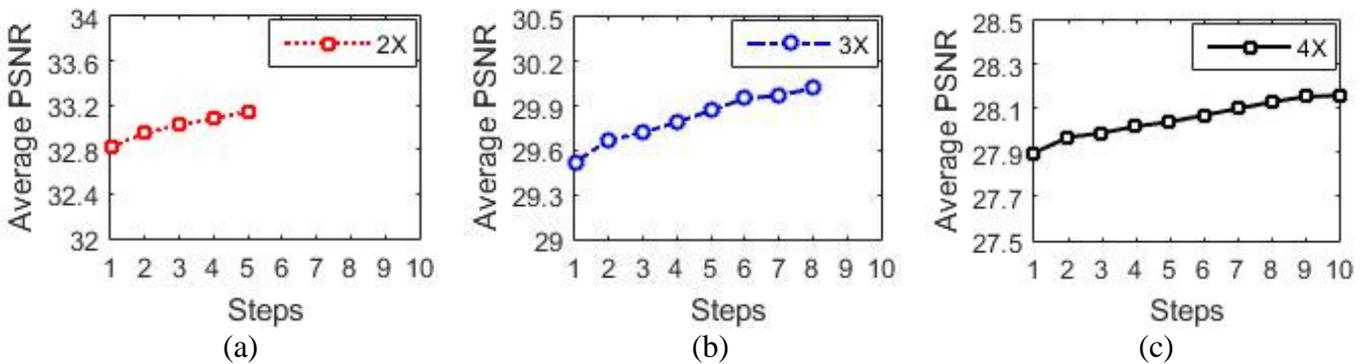

**FIGURE 4.** Average PSNR values with different number of steps on 'Set14', (a) 2 × magnification, (b) 3 × magnification, (c) 4 × magnification.

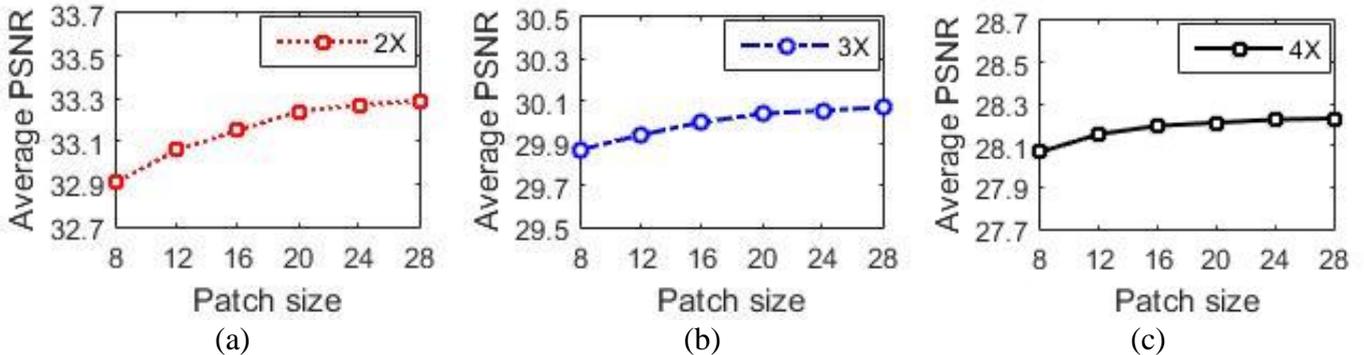

**FIGURE 5.** Average PSNR values with different patch sizes on 'Set14', (a) 2 × magnification, (b) 3 × magnification, (c) 4 × magnification.

## III. IMPLEMENTATION DETAILS

### A. SELECTION OF THE DEPTH

In [45], Kim *et al.* have proven that deeper network can obtain better SR results. Hence, it is important to select an appropriate depth of the network. In the GUN, the depth of the network is mainly decided by two parameters, *i.e.*, the number of upsampling steps, and the depth of the sub-network in each step. In the following, we test the selection of these two parameters separately.

### 1) SELECTION OF DEPTH IN EACH STEP

Fig.3 illustrates the average PSNR values on 'Set14' with different depths in each step. The total number of steps is fixed as 4, and the size of input patch is 12 × 12. Each step consists of several 3 × 3 conv. layers and one last 1 × 1 conv. layer. As shown in Fig.3, it can be found that the deeper network also performs the better. However, the PSNR results increase slowly when the depth is larger than 4, presumably because of the fixed training settings and finite training samples. The deeper network also costs more training and testing time. The depth in each step is thus set to 4 in the following experiment.

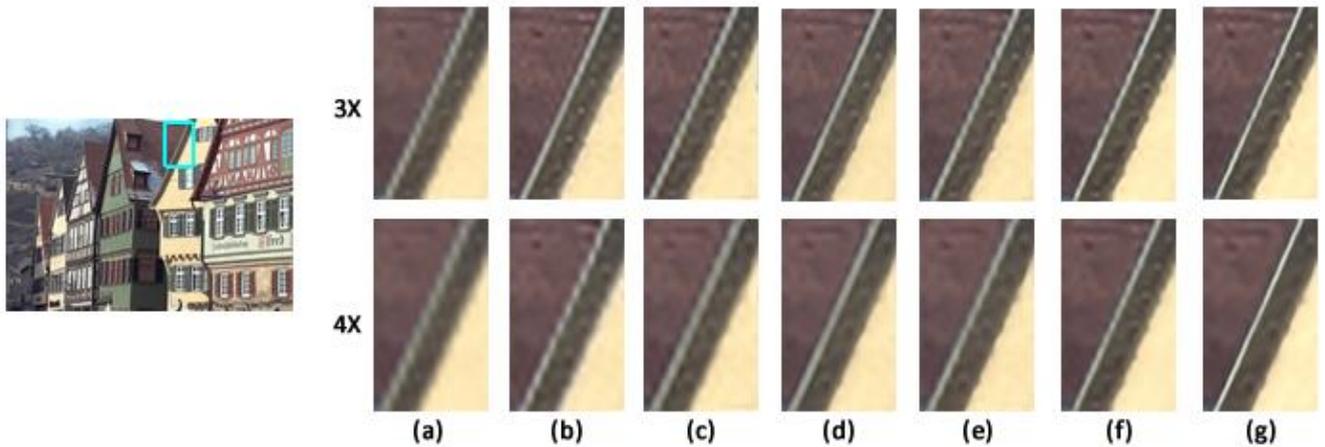

**FIGURE 6.** $3\times$ and $4\times$ upsampled results of 'kodim08' image with different methods, (a) with bicubic, (b) with the ASDS [32], (c) with the ANR [42], (d) with the A+ [43], (e) with the SRCNN [44], (f) with the VDSR [45], (g) with the GUN.

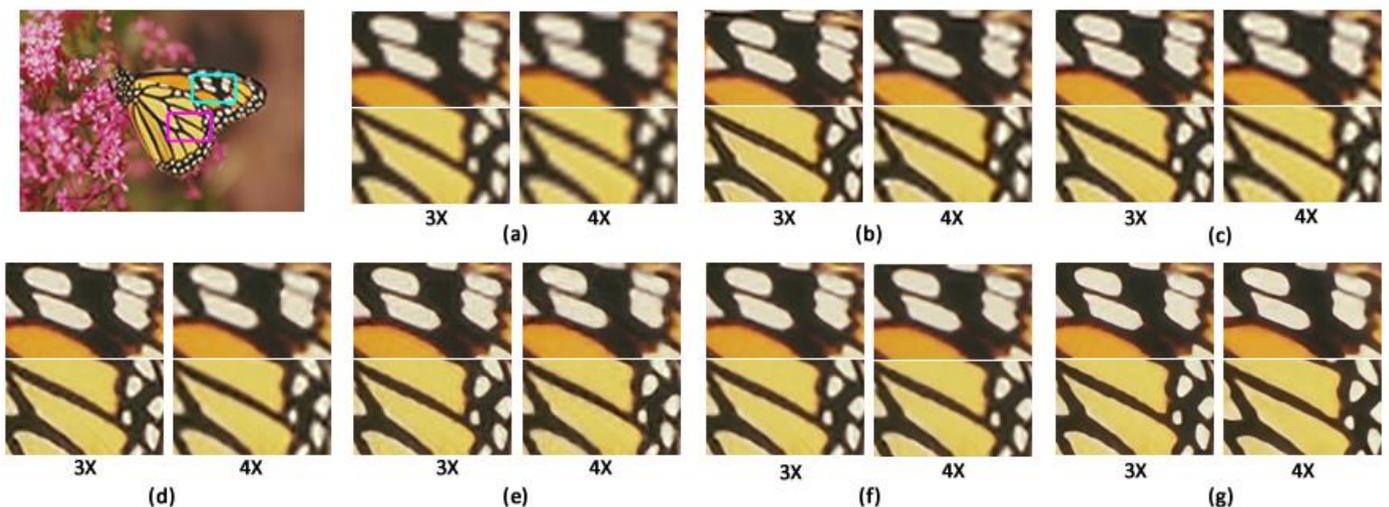

**FIGURE 7.** $3\times$ and $4\times$ upsampled results of 'monarch' image with different methods, (a) with bicubic, (b) with the ASDS [32], (c) with the ANR [42], (d) with the A+ [43], (e) with the SRCNN [44], (f) with the VDSR [45], (g) with the GUN.

### 2) SELECTION OF STEPS

Fig.4 shows the relationship between the average PSNR results and the total number of steps. The input patch size is also $12\times12$, and the depth in each step is fixed to 4. On the whole, increasing the number of steps can improve the performance. This also demonstrates the deeper the network the better the results. Note that the number of steps is also influenced by the increased resolution between the LR and HR training patches. In this paper, the numbers of steps are experimentally set as 5, 8, and 9 for $2\times$, $3\times$, and $4\times$ magnification, respectively.

### B. SIZE OF THE INPUT PATCH

The average PSNR values with different patch sizes are shown in Fig.5. We can find that enlarging the patch can also slightly increase the PSNR results. It should be noticed that the quantity of samples are fixed as 200, 000 to train the networks. In practice, larger patch size also leads to the reduction of total training samples. As a result, the patch sizes are selected by considering both the performance and the total number of training samples. In our experiment, the patch sizes are chosen as 20, 16, and 12 for $2\times$, $3\times$, and $4\times$ magnification, respectively.

## IV. EXPERIMENTAL RESULTS

### A. TESTING IMAGE SETS

For testing, three typical and largish image sets are used, *i.e.*, subset of 'Set14' [36], 'B100' [42], and Kodak PhotoCD dataset [4]. 'B100' selects 100 images from the Berkeley

---

[4] http://r0k.us/graphics/kodak/

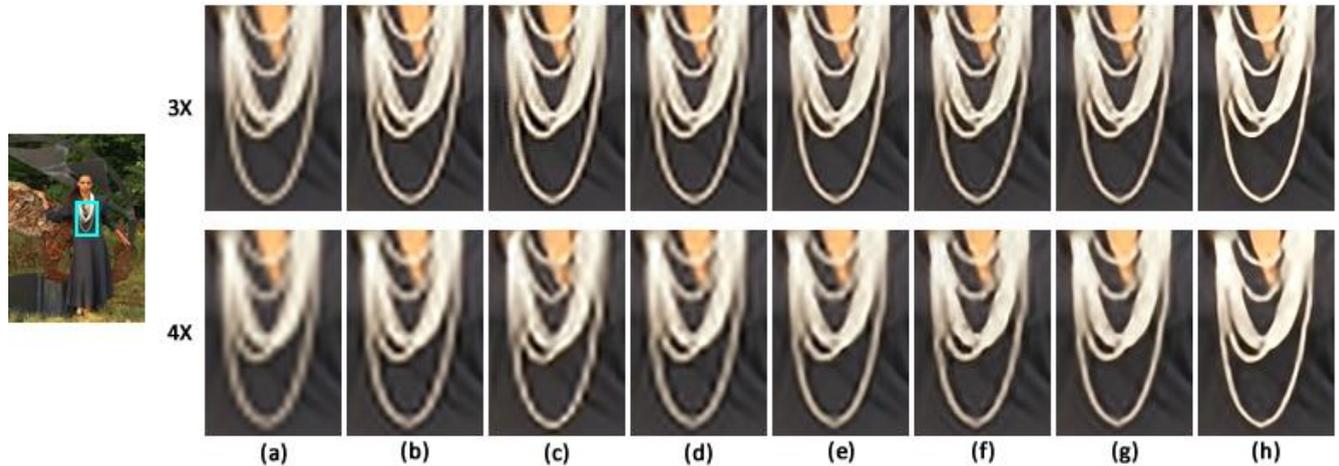

**FIGURE 8.** 3 × and 4 × upsampled results of 'kodim18' image with different methods, (a) with bicubic, (b) with the LLE [26], (c) with the ASDS [32], (d) with the ANR [42], (e) with the A+ [43], (f) with the SRCNN [44]. (g) with the VDSR [45], (h) with the GUN.

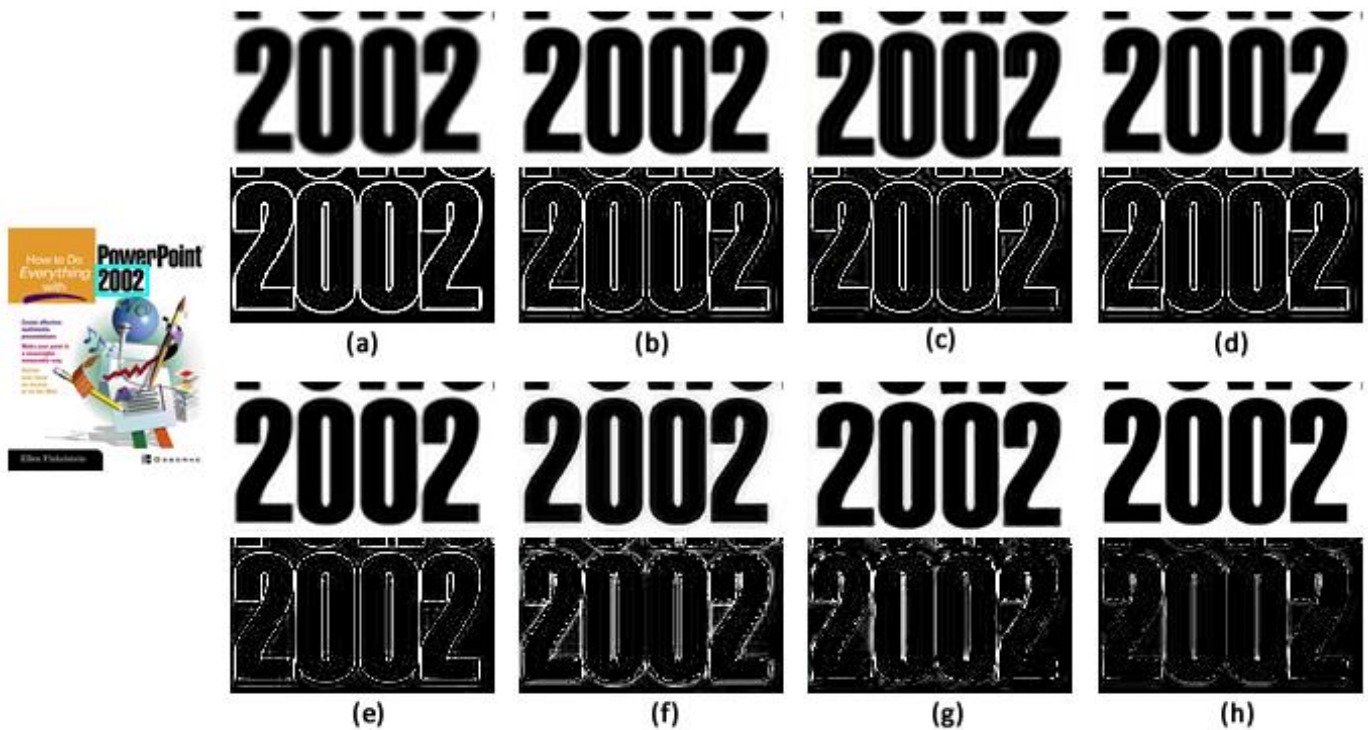

**FIGURE 9.** 2 × SR results of 'PPT' image with different methods, (a) with bicubic, (b) with the LLE [26], (c) with the ASDS [32], (d) with the ANR [42], (e) with the A+ [43], (f) with the SRCNN [44]. (g) with the VDSR [45], (h) with the GUN. The residual map between each result and the original HR image is illustrated.

Segmentation dataset (BSD) [54]. The Kodak PhotoCD dataset consists of 24 lossless true color images without compression artifacts, and is used as a standard testing set for many image processing works [66]. Note that 'Set14' even contains zero-padded 6-bit image (e.g. the 'bridge') [70], which is unreasonable to be used for 8-bit image SR experiment. Hence, 11 color images are selected from original 'Set14' to constitute a subset in our experiment.

## B. COMPARISON WITH OTHER METHODS

In this section, we compare the proposed GUN with some learning-based methods, such as the LLE [26], the ScSR [30], the ASDS [32], the ANR [42], the A+ [43], the SRCNN [44], and the VDSR [45]. In t experiment, the color testing images are firstly converted from RGB to YCrCb, and these SR methods are applied only on Y (intensity) component. The rest two channels are simply reconstructed with bicubic interpolation. In our experiment, the LR inputs are obtained by downsampling the original HR images with bicubic interpolation. Note that the ASDS firstly filtered and then downsampled the HR image to obtain the LR input, which is slightly different to other methods. The magnification factors in this paper are set as 2, 3, and 4.

Fig.6 compares the $3 \times$ and $4 \times$ magnified results with different SR methods. The zoom-up area is marked with blue rectangle, and the HR ground truth is also given. By comparing the upsampled edges of the roof, we can make the following observations. First, these learning-based methods can recover much better lines than bicubic. Second, the two CNN-based methods, *i.e.*, the VDSR and the GUN, can reproduce sharp edges for $4 \times$ SR. Last, by comparing the details around the roof edges, the proposed GUN obtains clearer and sharper results than other methods.

Fig.7 also illustrates the $3 \times$ and $4 \times$ upsampled results of 'monarch' image with different methods. Two area on the wings of the monarch are shown, from which we can get the following findings. First, the bicubic interpolated results are very jaggy and blurring, and the learning-based methods can reconstruct sharper edges. The results of the GUN are the sharpest among them. Second, the VDSR and the GUN can reproduce fine flat area than other methods, and that make their results much cleaner and clearer. Last, by comparing the tiny lines in the $4 \times$ results, the GUN can recover more natural and better details by means of the gradual learning and upsampling.

Fig.8 compares the $3 \times$ and $4 \times$ SR results on another image. We can get some findings similar to Fig.6. The A+, the SRCNN, the VDSR and the proposed GUN can reconstruct clear and smooth edges. By comparing the necklace area, the GUN still recovers sharper and clearer edges than other comparisons.

The $2 \times$ results on image 'PPT' with various methods are shown in Fig.9. We can find all these learning-based methods can recover fine edges for small magnification factor. To facilitate the comparison of subjective quality, the residual map between each result and the original HR image is multiplied by 20 and then illustrated. By comparing the details around the digits, we can obtain some observations. First, the result of bicubic interpolation is blurry but without fake edges. Second, the VDSR and the GUN can reproduce clear edges, while other methods suffer from either ringing effects or fake edges. Last, by comparing the residual components, the GUN achieves the least difference to the HR ground truth.

For objective quality assessment, we utilize two common evaluation metrics of the PSNR and the SSIM [63]. However, the information fidelity criterion (IFC) [65] is proved to have higher correlation with human ratings for SR evaluation than PSNR and SSIM [64]. Thus we also adopt the IFC in the experiment to estimate the subjective quality of different SR results. Table II, Table III, and Table IV list the objective assessment results on three datasets of the 'B100', the subset of 'Set14', and the Kodak dataset, respectively. From these tables, it can be found that the proposed GUN can achieve higher PSNR and SSIM values than other methods for different magnification factors and different datasets. By comparing the IFC values, the proposed GUN also obtains better IFC results than other methods. These results also demonstrate the effectiveness of the proposed network and the training processes. Note that the SRCNN and VDSR are also trained with Yang's 91 images [30] to make a fair comparison with traditional learning-based methods. Indeed, CNN-based methods can achieve much better results by means of larger training sets [47]. In Table II, we also list upsampling results of the SRCNN trained on subset of ImageNet [44], and the VDSR trained with 91 images and 200 BSD images [45]. It can be easily found that the CNN-based models benefit a lot from sufficient training data, and the upsampling results of the proposed GUN can also be further improved by means of more training images.

## C. FURTHER ANALYSES

Finally we again describe the benefits and further analyze the limitations of the proposed GUN. The gradual upsampling can reduce the difficulty of training a direct upsampling network. Furthermore, the proposed training process can rapidly train an initial network and then gradually optimize the weights. By comparing with the efficient deep network VDSR [45], although the residual-training and gradient clipping strategies are not used in the proposed network, the GUN is still converged very fast in the first several epochs. The GUN can be regarded as a variant of the VDSR which is specially designed for the SR scenario. But for other image reconstruction problems, such as denoising and deblocking, the VDSR is still much easier to be applied than the GUN, since the upsampling process maybe not needed in these tasks. Recently, especially in recent SR competition [52], many state-of-the-art ResNet-based SR methods have been proposed and outperform traditional CNN-based models by means of deeper network and skip connections. Since CNN is the basic model of SR network, the proposed gradual structure can naturally be applied for ResNet-based models.

## V. CONCLUSIONS

In this paper, an efficient deep convolutional neural network based super-resolution method has been proposed, namely Gradual Upsampling Network (GUN). The proposed GUN consists of an input layer, multistep upsampling and convolutional layers, and an output layer. The difficult direct upsampling problem is relaxed to several easier gradual upsampling processes with very small magnification factors. Hence, the GUN can efficiently learn to reconstruct the HR results, step-by-step. Furthermore, we present a gradual training process for the GUN, in which the simple edge-like patches are firstly utilized to train an initial network and then more complex patches are added to tuning the weights. Experimental results on three representative image datasets demonstrate that the proposed gradual structure and training strategy can promote the performance of conventional CNN-based method, and make the training phase much easier.

TABLE II.
AVERAGE PSNR (dB), SSIM, AND IFC OF DIFFERENT METHODS ON IMAGE SET "B100"

| | Training Set | 2X | | | 3X | | | 4X | | |
|---|---|---|---|---|---|---|---|---|---|---|
| | | PSNR | SSIM | IFC | PSNR | SSIM | IFC | PSNR | SSIM | IFC |
| Bicubic | [30] | 29.35 | 0.8334 | 5.85 | 27.17 | 0.7361 | 3.47 | 25.95 | 0.6671 | 2.29 |
| LLE | [30] | 30.40 | 0.8674 | 6.12 | 27.84 | 0.7687 | 3.95 | 26.47 | 0.6937 | 2.74 |
| ScSR | [30] | 30.32 | 0.8709 | 6.24 | 27.74 | 0.7719 | 4.22 | 26.33 | 0.6997 | 2.85 |
| ASDS | [30] | 30.19 | 0.8712 | 6.72 | 27.65 | 0.7735 | 4.24 | 26.25 | 0.7003 | 2.97 |
| Zeyde | [30] | 30.40 | 0.8682 | 6.32 | 27.87 | 0.7693 | 4.18 | 26.51 | 0.6963 | 2.76 |
| ANR | [30] | 30.50 | 0.8706 | 6.59 | 27.90 | 0.7724 | 4.16 | 26.52 | 0.6991 | 2.67 |
| A+ | [30] | 30.76 | 0.8762 | 6.64 | 28.18 | 0.7764 | 4.19 | 26.76 | 0.7062 | 2.72 |
| SRCNN | [30] | 31.06 | 0.8854 | 6.62 | 28.18 | 0.7780 | 4.14 | 26.79 | 0.7059 | 2.69 |
| VDSR | [30] | 31.30 | 0.8861 | 6.79 | 28.31 | 0.7789 | 4.24 | 26.93 | 0.7070 | 2.83 |
| GUN | [30] | 31.49 | 0.8889 | 6.96 | 28.49 | 0.7886 | 4.32 | 27.15 | 0.7191 | 3.06 |
| SRCNN[44] | [44] | 31.36 | 0.8879 | **7.24** | 28.41 | 0.7863 | - | 26.90 | 0.7101 | 2.41 |
| VDSR[45] | [45] | 31.90 | 0.8960 | 7.17 | 28.82 | 0.7976 | - | 27.29 | 0.7251 | 2.63 |
| GUN | [45] | **31.98** | **0.8977** | 7.20 | **28.94** | **0.7991** | **4.47** | **27.41** | **0.7274** | **3.09** |

TABLE III.
AVERAGE PSNR (dB), SSIM, AND IFC OF DIFFERENT METHODS ON "SUBSET OF SET14"

| | Training Set | 2X | | | 3X | | | 4X | | |
|---|---|---|---|---|---|---|---|---|---|---|
| | | PSNR | SSIM | IFC | PSNR | SSIM | IFC | PSNR | SSIM | IFC |
| Bicubic | [30] | 30.36 | 0.9417 | 5.83 | 27.67 | 0.8596 | 3.41 | 26.12 | 0.7857 | 2.27 |
| LLE | [30] | 31.91 | 0.9587 | 6.08 | 28.74 | 0.8836 | 3.89 | 26.95 | 0.8137 | 2.21 |
| ScSR | [30] | 31.21 | 0.9620 | 6.22 | 28.01 | 0.8882 | 4.04 | 26.57 | 0.8183 | 2.65 |
| ASDS | [30] | 31.15 | 0.9627 | 6.61 | 27.91 | 0.8938 | 4.11 | 26.94 | 0.8190 | 2.35 |
| Zeyde | [30] | 31.96 | 0.9589 | 6.25 | 28.80 | 0.8841 | 4.02 | 26.99 | 0.8159 | 2.67 |
| ANR | [30] | 31.95 | 0.9626 | 6.36 | 28.80 | 0.8890 | 3.67 | 27.00 | 0.8194 | 2.48 |
| A+ | [30] | 32.39 | 0.9641 | 6.54 | 29.12 | 0.8940 | 4.04 | 27.34 | 0.8294 | 2.62 |
| SRCNN | [30] | 32.93 | 0.9648 | 7.06 | 29.54 | 0.9023 | 3.94 | 27.85 | 0.8458 | 2.72 |
| VDSR | [30] | 33.07 | 0.9689 | 7.16 | 29.82 | 0.9055 | 4.17 | 28.07 | 0.8533 | 2.75 |
| GUN | [30] | **33.35** | **0.9698** | **7.31** | **30.08** | **0.9112** | **4.31** | **28.29** | **0.8648** | **2.96** |

TABLE IV.
AVERAGE PSNR (dB), SSIM, AND IFC OF DIFFERENT METHODS ON IMAGE SET "KODAK"

| | Training Set | 2X | | | 3X | | | 4X | | |
|---|---|---|---|---|---|---|---|---|---|---|
| | | PSNR | SSIM | IFC | PSNR | SSIM | IFC | PSNR | SSIM | IFC |
| Bicubic | [30] | 30.88 | 0.9787 | 5.46 | 28.46 | 0.8996 | 3.23 | 27.26 | 0.8438 | 2.13 |
| NE+LLE | [30] | 32.22 | 0.9937 | 6.81 | 29.21 | 0.9224 | 3.77 | 27.80 | 0.8711 | 2.45 |
| ScSR | [30] | 32.15 | 0.9923 | 6.65 | 29.07 | 0.9115 | 3.58 | 27.69 | 0.8707 | 2.32 |
| ASDS | [30] | 32.13 | 0.9930 | 6.79 | 29.11 | 0.9202 | 3.72 | 27.79 | 0.8715 | 2.51 |
| ANR | [30] | 32.28 | 0.9939 | 6.93 | 29.25 | 0.9233 | 3.82 | 27.84 | 0.8724 | 2.46 |
| A+ | [30] | 32.75 | 0.9949 | 7.12 | 29.61 | 0.9278 | 3.90 | 28.13 | 0.8789 | 2.50 |
| SRCNN | [30] | 32.74 | 0.9934 | 7.09 | 29.39 | 0.9274 | 3.86 | 28.14 | 0.8777 | 2.42 |
| VDSR | [30] | 32.80 | 0.9950 | 7.19 | 29.59 | 0.9285 | 3.95 | 28.21 | 0.8790 | 2.53 |
| GUN | [30] | **33.09** | **0.9959** | **7.27** | **29.73** | **0.9293** | **4.04** | **28.36** | **0.8811** | **2.76** |

## ACKNOWLEDGEMENTS

The authors would like to sincerely thank the anonymous reviewers. We also sincerely thank R. Timofte, and C. Dong for sharing the source codes of the ANR/A+, and the SRCNN methods. The authors also thank Dr. Jianchao Yang, Dr. Wangmeng Zuo, Dr. Weisheng Dong, and Dr. Jinshui Hu for useful advice during this work.